\pdfoutput=1

\documentclass[11pt]{article}

\usepackage{EMNLP2023}

\usepackage{times}
\usepackage{latexsym}

\usepackage[T1]{fontenc}

\usepackage[utf8]{inputenc}

\usepackage{microtype}

\usepackage{inconsolata}
\usepackage{multirow}
\usepackage{tcolorbox}
\usepackage{graphicx}
\usepackage{stfloats}
\usepackage{caption}
\usepackage[font=small]{caption}
\title{Autonomous Deep Agent}

\author{Amy Yu \and Erik Lebedev \and Lincoln Everett \and  Xiaoxin Chen \and Terry Chen \\[10pt]
Deep Agent Team \\[5pt]
\small \texttt{\{amy.yu, erik.lebedev, lincoln.everett, xiaoxin.chen, terry.chen\}@theagentix.ai} \\
\small \texttt{\{deepagent\}@theagentix.ai}
}

\begin{document}
\maketitle
\begin{abstract}
This technical brief introduces Deep Agent, an advanced autonomous AI system designed to manage complex multi-phase tasks through a novel hierarchical task management architecture. The system's foundation is built on our Hierarchical Task DAG (HTDAG) framework, which dynamically decomposes high-level objectives into manageable sub-tasks while rigorously maintaining dependencies and execution coherence. Deep Agent advances beyond traditional agent systems through three key innovations: First, it implements a recursive two-stage planner-executor architecture that enables continuous task refinement and adaptation as circumstances change. Second, it features an Autonomous API \& Tool Creation (AATC) system that automatically generates reusable components from UI interactions, substantially reducing operational costs for similar tasks. Third, it incorporates Prompt Tweaking Engine and Autonomous Prompt Feedback Learning components that optimize Large Language Model prompts for specific scenarios, enhancing both inference accuracy and operational stability. These components are integrated to form a service infrastructure that manages user contexts, handles complex task dependencies, and orchestrates end-to-end agentic workflow execution. Through this sophisticated architecture, Deep Agent establishes a novel paradigm in self-governing AI systems, demonstrating robust capability to independently handle intricate, multi-step tasks while maintaining consistent efficiency and reliability through continuous self-optimization.

\end{abstract}

\section{Introduction}
Modern artificial intelligence systems increasingly confront complex, multi-phase tasks that demand the orchestration of business workflows, in-depth analyses, and changing demands and contexts. These complex tasks feature dynamic dependencies where subsequent steps could rely on earlier outcomes, with requirements that evolve in real time. As the Large Language Models (LLM) become more powerful, we witness a decisive shift from single-step AI solutions toward autonomous deep agents capable of planning, executing, and adapting throughout entire processes \cite{masterman2024landscape}. The most recent advances include OpenAI's DeepResearch \cite{openai2025deepresearch} and Operator \cite{openai2025operator}. Through the integration of multi-step logical reasoning, external memory architectures, and sophisticated tool utilization capabilities, the next generation of AI agents is positioned to revolutionize end-to-end process automation \cite{wsj2025operator}. This evolution toward autonomous agents that can reason and act on behalf of users—managing task coordination, handling complex dependencies, and responding to new information—represents a significant frontier in AI capabilities.

In this technical brief, we introduce \textbf{\textit{Deep Agent}}, our novel LLM-driven AI framework designed to address these challenges and advance autonomous task execution and management. 

Deep Agent models complex tasks using a \textbf{Hierarchical Task DAG (HTDAG)}, which recursively represents sub-tasks and their dependencies across multiple layers of a Directed Acyclic Graph (DAG). This DAG-based design enables a dynamic decomposition of high-level goals into a network of interrelated sub-tasks (nodes in the graph), with directed edges capturing their dependencies and constraints. Unlike static workflows, our HTDAG facilitates \textbf{dynamic task management}, allowing for the expansion of nodes into finer-grained sub-tasks as needed and modification of the graph in real time as new information becomes available or objectives change. DAG naturally supports user chiming in anytime to pause, terminate, resume or adjust the task, such as "let me review first after you found the recipes". This ensures flexibility and effectiveness in environments with evolving requirements or unforeseen events, addressing a critical limitation of static pipelines. Furthermore, \textbf{the hierarchical design enables disruption containment to support user co-piloting and failure recovery scenarios}. When users directly interface with tasks or when technical complications arise in complex operations, the hierarchical structure localizes potential disruptions to specific levels. This containment prevents cascading effects across the broader task graph, enhancing the system's overall resilience and reliability.

Deep Agent supports \textbf{context-aware and user-aware decision-making}; when selecting which task to execute next or determining the method of execution, the agent considers the current state of the workflow, results from previous actions, and any user context and feedback. This ensures that decisions are informed by the history and broader context of the task sequence, facilitating proactive interactions with users for clarification and disambiguation.

Deep Agent executes tasks through a comprehensive framework comprising actions, APIs, and tools. \textit{Actions} are direct operations the agent performs on user interfaces, such as clicking buttons, selecting options, or entering text on websites or mobile applications. \textit{APIs} provide direct access to specific service functionalities, while \textit{Tools} are higher-level constructs that combine multiple actions and APIs to achieve more high-level objectives. Unlike previous agent systems that rely only on a pre-defined set of APIs and tools, Deep Agent autonomously develops new APIs and tools by leveraging existing UI actions and APIs. This innovative approach, termed \textbf{Autonomous API \& Tool Creation (AATC)}, enhances the reusability of previous LLM inference results and allows for the pre-processing of various UIs. Consequently, it significantly reduces LLM inference overhead when handling similar tasks, leading to lower marginal costs and improved industrial applicability for the agent system.

Deep Agent is a generic agent system utilizing general-purpose LLMs as its backend and generic prompts; however, it recognizes that specific mindset adjustments may be necessary for different tasks. A \textbf{Prompt Tweaking Engine (PTE)} is developed to modify existing instructions or append additional instructions to the generic prompts for various scenarios. This technique reduces unnecessary instructions when they are not needed, therefore decreasing prompt size and complexity, and significantly improving LLM inference accuracy and stability with a few more inferences before actually addressing the task \footnote{"a few more inference steps" is considered acceptable in comparison to dozens of inference steps needed for complex task's end-to-end workflows.}. This approach aligns with prompting methods that dynamically optimize the prompt before the actual inference to enhance LLM performance \cite{chang2024efficient}. This technique is applied to the aforementioned AATC component to pre-process UIs to offset runtime latency.

Lastly but not the least, Deep Agent incorporates closed-loop \textbf{Autonomous Feedback Learning (AFL)} at the core of its design. Techniques such as reflections are employed to detect mistakes and enhance performance. Disagreements from the reflections are heavily post-examined to identify agent flaws and potential improvements. The feedback learning includes leveraging Deep Agent itself as the evaluator framework (look at the same task agent results from an evaluator's perspective), autonomously developing prompt tweaks for specific scenarios for the PTE component to leverage, as well as providing feedback to the model's pre- and post-training data. Deep Agent is thus ensured with continuous improvement and adaptation. 

By combining these elements, our Deep Agent represents a significant step toward industry viable autonomous AI systems that are self-directed and resilient in managing complx tasks and their entire complex end-to-end workflows.

\section{Framework Design}

\subsection{Hierarchical Task DAG}

\begin{figure*}[h!]
    \centering
    \includegraphics[width=1.0\textwidth]{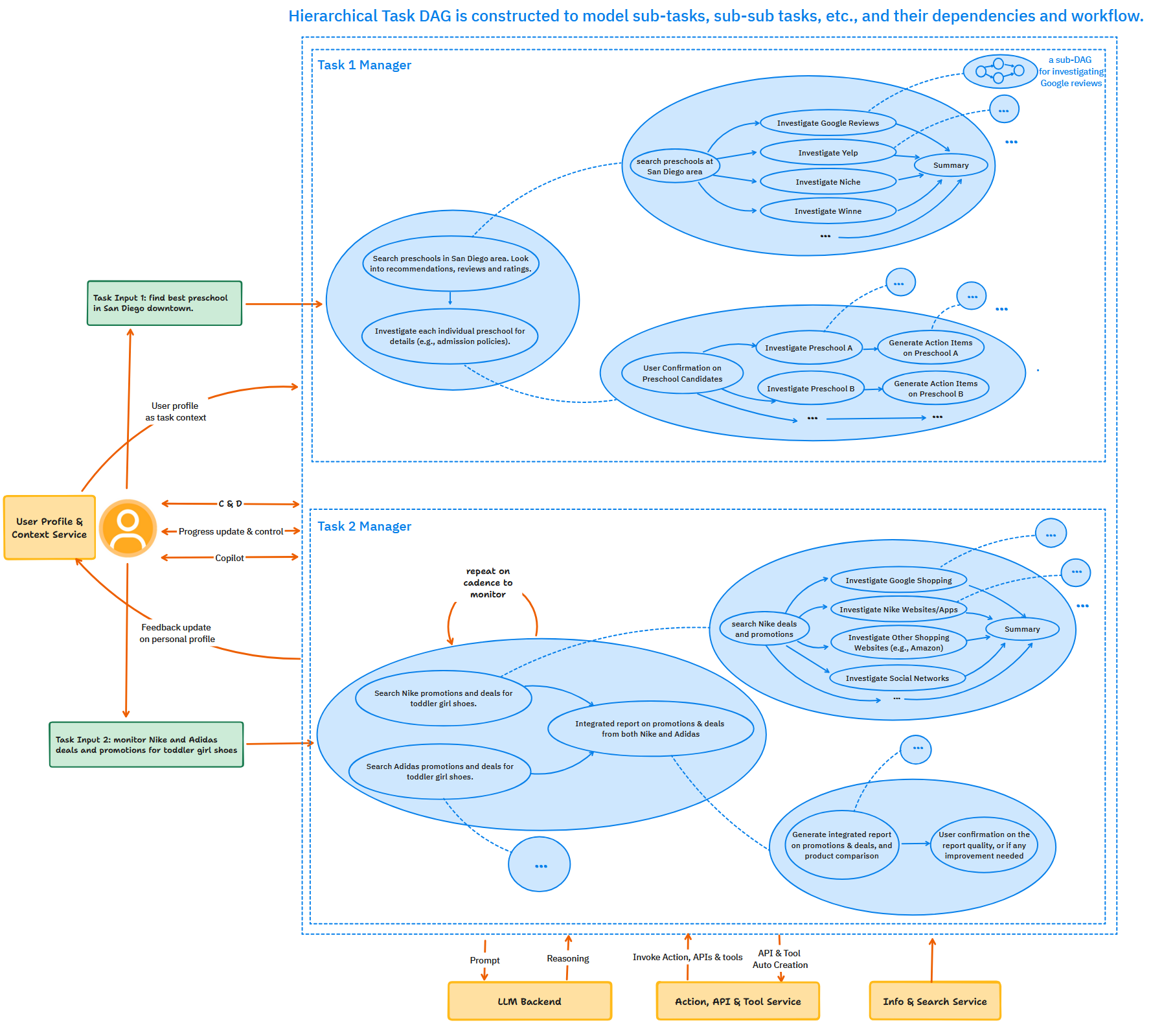} 
    \caption{Deep Agent system overview illustrating hierarchical task decomposition and workflow management. The system demonstrates two representative tasks: finding the best preschool (Task 1) and monitoring shopping deals (Task 2). Each task is managed by a dedicated Task Manager that constructs Hierarchical Task DAGs (HTDAGs) to model sub-tasks and their dependencies. The system integrates with users through a ``User Profile \& Context Service'' that enables personalization and runtime interactions (e.g., control and dynamic feedback, progress updates, and copilot functionality). Task execution is facilitated through three core services: an ``LLM Backend'' for reasoning, an ``Info \& Search Service'' for information gathering, and an ``Action, API \& Tool Service'' for end execution. The system implements two key feedback mechanisms: 1) runtime learning of user preferences for profile enhancement, and 2) autonomous creation of new APIs and tools to expand system capabilities.}
    \label{fig:HTDAG}
\end{figure*}

\begin{figure*}[h!] 
    \centering
    \includegraphics[width=1.0\textwidth]{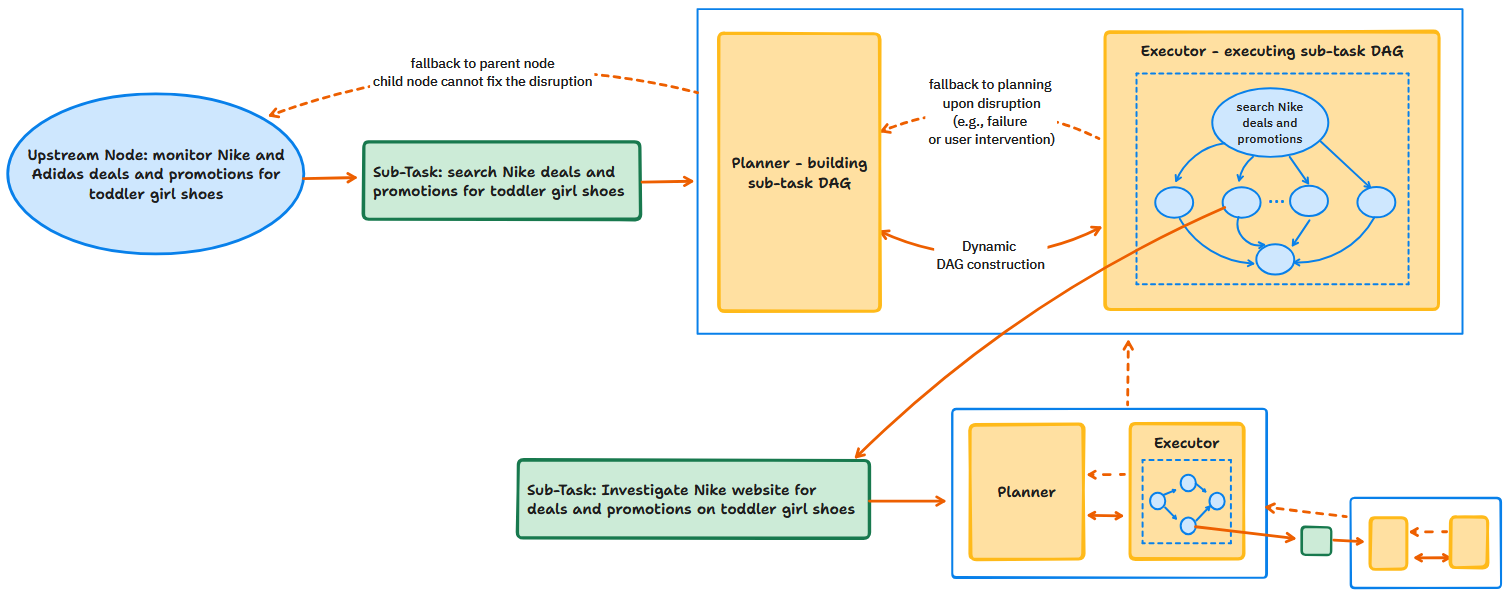} 
    \caption{Two-stage planner-executor architecture illustrated through a Nike deals monitoring example. This architecture decomposes a high-level task into sub-tasks, each through a dedicated planner-executor pipeline, enabling recursive task refinement, as shown by the sub-task ``Investigate Nike website ...'' being further decomposed through another planner-executor cycle. At each cycle, the planner dynamically constructs sub-task DAGs for workflow to the best given all available information at the moment (see also Figure \ref{fig:HTDAG_dynamics}), while the executor manages tactical execution. Upon failures or disruptions (such as additional constrains from user, or direct user intervention), re-planning is triggered to consider both execution state and the disruptions. Unresolved disruptions cascade to parent nodes for higher-level re-planning. This design allows sophisticated task management and resilient handling of complex, nested workflows.}
    \label{fig:HTDAG_two_stage}
\end{figure*}

Our framework employs a Hierarchical Task DAG (HTDAG) to model complex task workflows and their dependencies. While DAGs have long been utilized in computer science for task scheduling and dependency management, our hierarchical adaptation provides unique advantages for autonomous agent systems. Figure \ref{fig:HTDAG} presents a high-level overview of Deep Agent, highlighting HTDAG as the principal component, while Figure \ref{fig:HTDAG_two_stage} illustrates our two-stage planner-executor architecture for each node.

HTDAG in general operates through a recursive two-stage \textit{planner-executor cycles} for each break down, as demonstrated in Figure \ref{fig:HTDAG_two_stage}. The \textit{planner} component dynamically constructs a next-level sub-task DAG given all available information at the moment, while the \textit{executor} component manages the execution of these task DAGs. When a downstream node receives a task from an up-stream node (not necessarily the original task; could be a sub-task) such as ``monitor Nike and Adidas deals and promotions for toddler girl shoes,'' the system first breaks it down into manageable next-level sub-tasks. A sub-task could be an \textit{atomic sub-task} that cannot be further decomposed (e.g., "click on button A") or a \textit{non-automic sub-task} (e.g., "investigate Nike website for deals and promotions on toddler girl shoes"), with non-atomic tasks triggering recursive planner-executor cycles for further decomposition.

\begin{figure*}[h!] 
    \centering
    \includegraphics[width=1.0\textwidth]{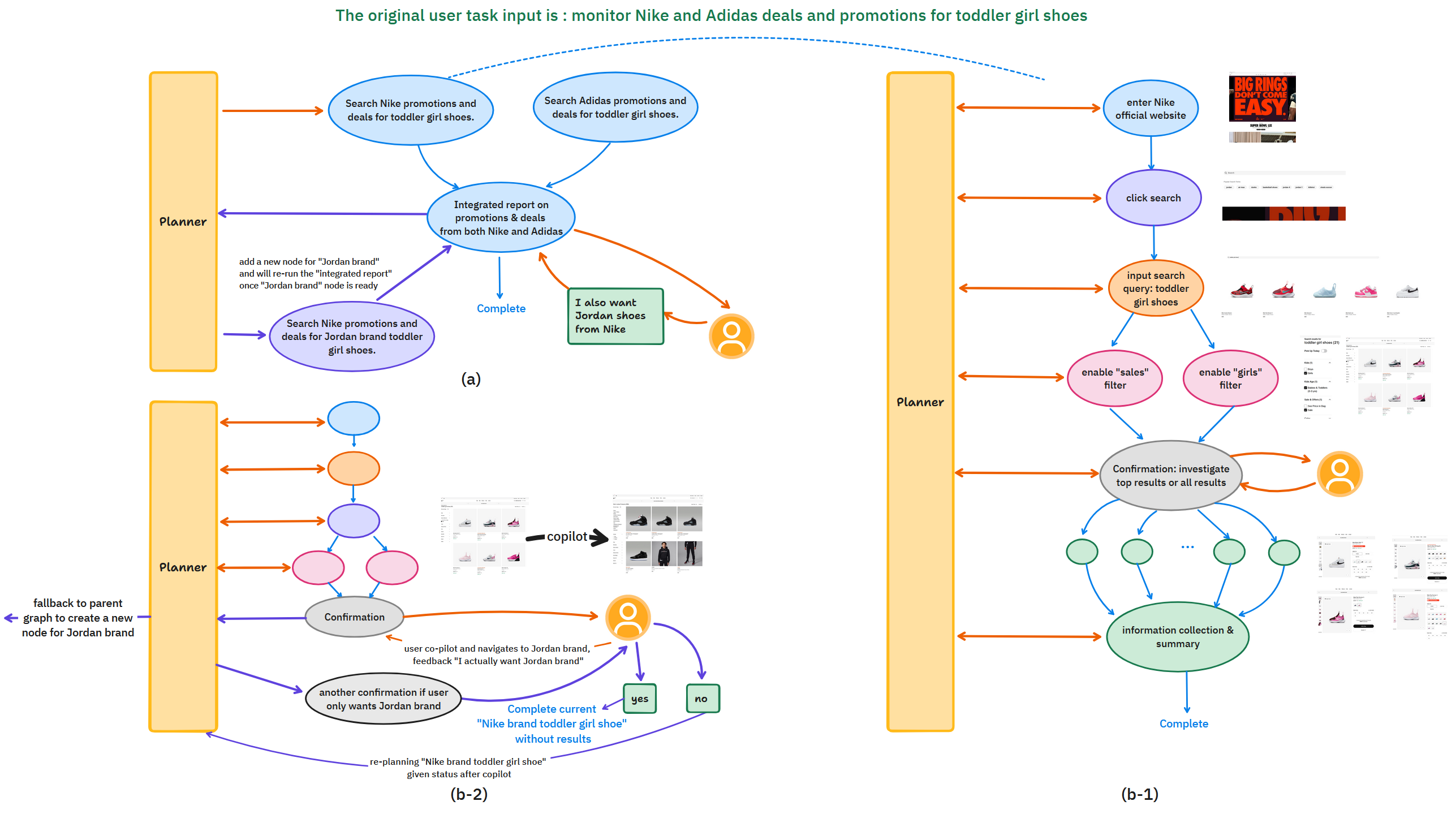} 
   \caption{\small Examples of dynamic DAG construction and adaptation during task execution. (a) Planner initially generated a plan to "search Nike, ...", "search Adidas ...", and then generate an integrated report on the promotions, deals and product comparison. During the report generation, user input (``I also want Jordan shoes'') triggers graph expansion to create a new task node responsible for "the Jordan brand" specifically and will re-run the "integrated report" node, while preserving all existing "search Nike, ...", "search Adidas ..." nodes. (b-1) For the thread of investigating Nike brand promotions \& deals, the task DAG construction is dynamic as new UI elements become available through the process. (b-2) If user copilot interaction happens during the process (assumed at the confirmation step in this diagram), and signaled "I actually wanted Jordan brand", re-planning is triggered, and it not only impacts the current DAG (confirm with user again, and either 1) terminate the current DAG if they only wanted Jordan brand, or 2) re-plans the current DAG given the latest status after copilot), but also falls back to the parent graph to create a new node for the "Jordan brand" task.}
    \label{fig:HTDAG_dynamics}
\end{figure*}

At each planner invocation, the task DAG is created/modified/expanded given the available information and context at the moment. This is esepcially important when doing ground work, such as when UI elements dyanmically change or appear. This is also important to support user intervention, such as initiating a new requirement in the process, or directly hands-on co-pilot the UI. The planner needs to be invoke to make appropirate adjustment to the task DAG, not just the current level, but could also be parent level, to adapt to changing needs, as shown in Figure \ref{fig:HTDAG_dynamics}.

The dynamic nature of DAG construction is crucial at each planner invocation. The task DAG is created, modified, or expanded based on the latest available information and context. This dynamic construction is particularly important in two scenarios: 1) during ground work where new information will dynamically change or appear (such as UI elements), and 2) during user interventions such as introducing new requirements or directly engaging in UI co-piloting. As shown in Figure \ref{fig:HTDAG_dynamics}, the planner makes appropriate adjustments to the task DAG not only at the current level but potentially at parent levels to adapt to changing needs.

Our HTDAG design provides the following key benefits:

\begin{itemize}
    \item \textbf{Adaptive Task Decomposition \& Refinement:} The planner creates next-level sub-task DAGs only when necessary, based on current context and requirements. This dynamic decomposition prevents the system from committing to overly detailed plans prematurely and getting lost, enabling effective handling of complex scenarios while maintaining LLM focus on immediately relevant tasks.
    
    \item \textbf{Coherent User Intervention Support:} The hierarchical structure naturally accommodates user interventions at different abstraction levels, while dynamic planning ensures these interventions are properly integrated into the workflow. When users provide new requirements or directly engage in co-piloting, the system can adjust both current and parent-level DAGs to maintain task coherence.
    
    \item \textbf{Failure Containment \& Recovery:} The hierarchical architecture localizes failures within specific levels, preventing cascade effects throughout the task graph. Combined with dynamic planning capabilities, this enables sophisticated recovery strategies where the system can re-plan failed components while preserving successfully executed parts of the task graph.
\end{itemize}

Beyond the hierarchical benefits, the design inherits several classical advantages of DAG-based architectures. The DAG structure naturally supports both sequential execution, which is crucial for scientific development and systematic debugging, and parallel execution for runtime optimization. Additionally, the DAG representation enables efficient dependency tracking, automated resource allocation, and incremental computation. The directed edges explicitly capture both data and control flow dependencies, facilitating state preservation and restoration, and sophisticated debugging/execution strategies such as partial re-execution. These benefits are particularly valuable for Deep Agent development, as they allow runtime optimization for complex task execution while maintaining clear execution traces for offline verification, analysis and debugging.

\subsection{Auto API \& Tool Creation}
\label{sec:aatc}

Deep Agent incorporates a novel component for Autonomous API \& Tool Creation (AATC). Unlike traditional LLM-driven systems that rely only on pre-defined API/tool sets, our framework can autonomously analyze, design, and build new reusable APIs and tools on top of UI interactions, and existing APIs \& tools). Once created, these APIs and tools become part of the system's permanent capabilities, significantly reducing the marginal cost of handling similar tasks by eliminating the need for repeated LLM inference on the UI interaction flow again.

The AATC leverages the core Deep Agent framework with specialized prompting for API and tool generation, operating through a systematic closed-loop process. Given a target UI (e.g., Nike's official website), the system first employs the Task Simulator to generate and execute diverse user scenarios at scale. Deep Agent then analyzes these interactions in conjunction with existing APIs and tools to identify gaps in coverage and opportunities for new capability creation. This analysis involves three key steps:

\begin{itemize}
    \item Identification of core UI functionalities and their interdependencies through systematic interaction analysis
    \item Extraction of fundamental user interaction patterns and common workflow sequences
    \item Specification of input/output requirements and parameter constraints for each identified functionality
\end{itemize}

\begin{figure*}[h!] 
    \centering
    \includegraphics[width=1.0\textwidth]{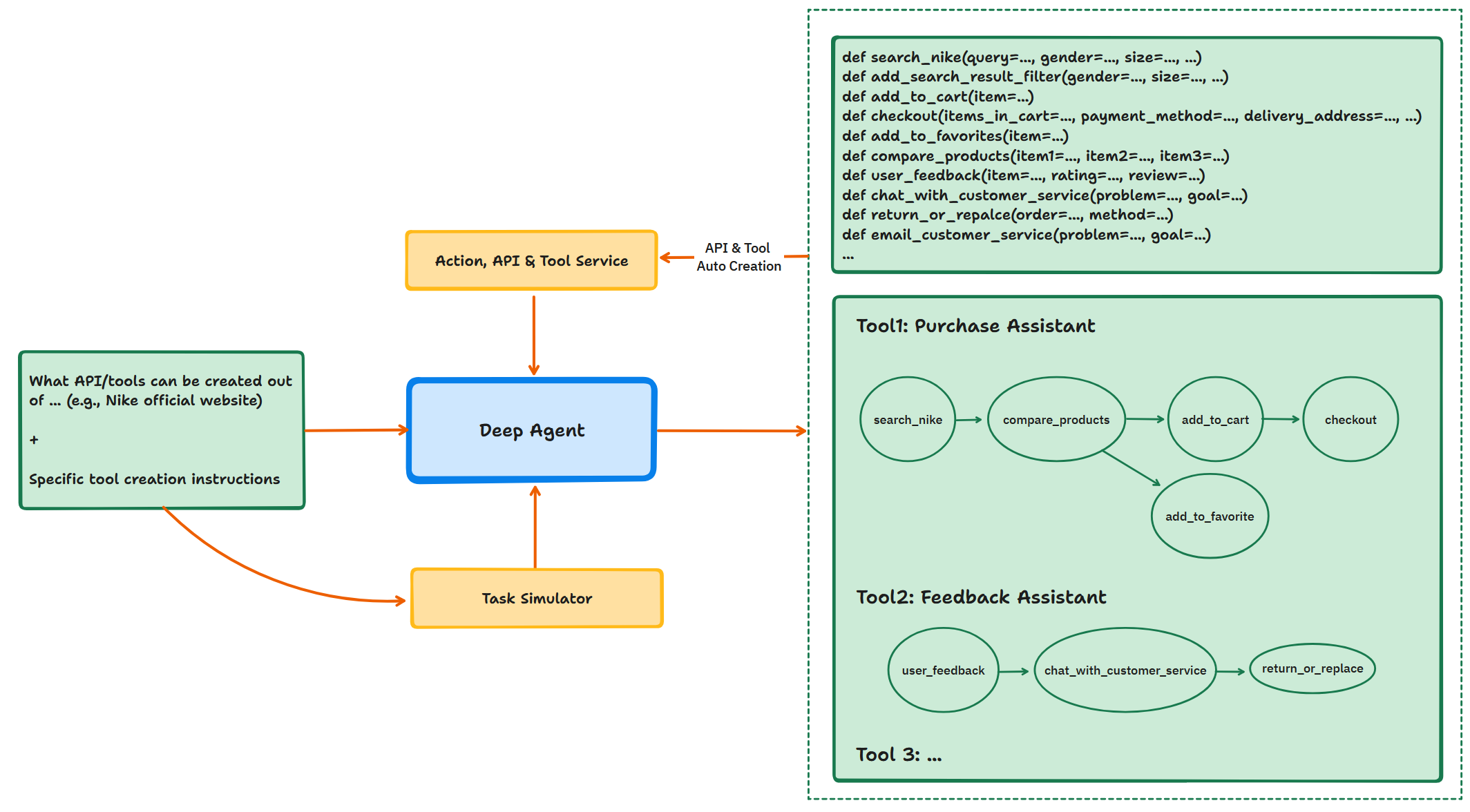} 
    \caption{Autonomous API \& Tool Creation (AATC) framework and examples. The system analyzes target UIs through Deep Agent and Task Simulator to automatically generate new APIs and composite tools. Left: The framework's closed-loop process where Deep Agent analyzes UI functionalities and creates new APIs/tools, which are then verified through task simulation. Right: Example outputs including automatically generated APIs (top) covering core UI functionalities with proper parameter specifications, and synthesized composite tools (bottom) that chain APIs into higher-level workflows. Shown examples include a Purchase Assistant tool that coordinates product search, comparison, and checkout operations, and a Feedback Assistant tool that manages user feedback and customer service interactions. This autonomous creation process enables continuous and scalable expansion of the system's capabilities through UI analysis.}
    \label{fig:attc}
\end{figure*}

Based on this analysis, AATC autonomously generates two levels of reusable components:
1) atomic APIs that encapsulate individual UI operations (e.g., search, filter, add to cart), and 
2) composite tools that chain multiple APIs into higher-level task workflows (e.g., Purchase Assistant, Feedback Assistant).
As shown in Figure \ref{fig:attc}, these components are automatically integrated into the ``Action, API \& Tool Service'', enabling reuse across similar tasks and contexts.

AATC also employs a continuous improvement cycle:
\begin{itemize}
    \item The API \& Tool service is enriched with the created APIs and tools.
    \item More simulation of potential user tasks, and Deep Agent continues to identify if even more new APIs and/or tools are possible.
\end{itemize}

This approach significantly reduces operational costs in two ways: by eliminating redundant LLM inference for common UI interactions, and by automatically expanding the system's reusable capability set through autonomous component creation. Figure \ref{fig:attc} demonstrates this through examples of both atomic APIs with well-defined interfaces and composite tools that encapsulate complex workflow patterns, all automatically generated from UI analysis.

\subsection{Prompt Tweaking Engine}
\label{sec:prompt_tweaking_engine}

A significant challenge in developing generic AI agents is balancing prompt design between generality and specificity. Deep Agent strives to use a minimal set of generic prompts, while still effectively handling diverse scenarios that may require specific rules. For instance, the rule of "prioritizing location verification" is critical for ordering from chain restaurant (e.g., Starbucks) and instant delivery platforms (e.g., Uber Eats, DoorDash) where store and service availability are location-dependent. In contrast, these rules are less relevant for national e-commerce platforms like Amazon or Nike, where delivery address verification can be deferred until checkout. However, such specialized rules tend to accumulate over time, particularly through the autonomous feedback learning process (Section \ref{sec:auto_feedback_learning}) that addresses edge cases by introducing additional rules. The resulting expansion of prompt complexity can lead to degraded LLM performance.

Deep Agent addresses this challenge through its Prompt Tweaking Engine (PTE), which dynamically optimizes prompts before task execution. While maintaining only a few base prompts containing both generic and specific instructions, PTE introduces a pre-processing step to filter out irrelevant rules based on the task context. This optimization process adds overhead typically under a dozen seconds even when employing techniques detailed in Section \ref{sec:test_time_computation_reflection_and_validation} — which is not negligible but reasonable compared to typical agent workflow durations of several minutes (e.g., 2-3 minutes for order completion). The resulting reduction in prompt size improves later inference efficiency, effectively offsetting the initial optimization cost. For well-defined scenarios where rule filtering can be reliably performed through pattern matching, PTE employs contextual embedding retrieval to further minimize latency impact.

\begin{figure}[h!]
    \centering
    \includegraphics[width=\columnwidth]{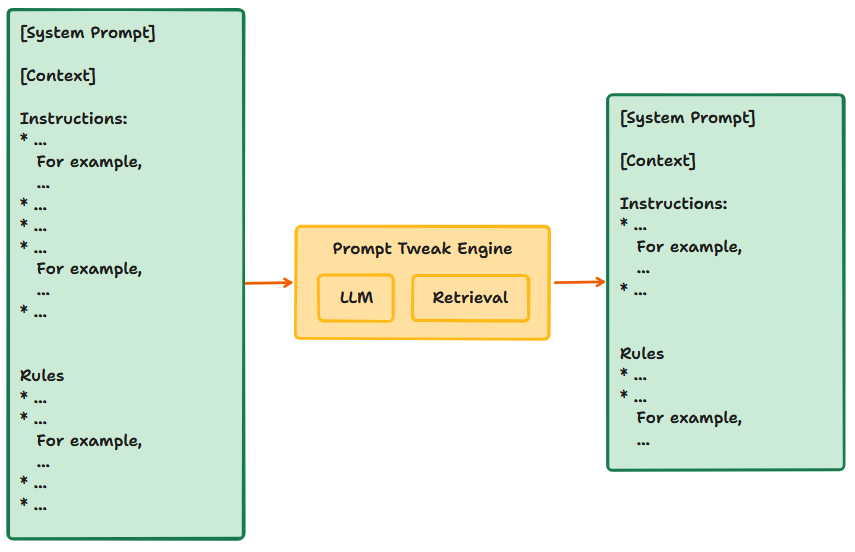}
    \caption{\small The Prompt Tweaking Engine to reduce irrelevant instructions and rules. The engine processes input prompts containing generic instructions and rules (left), applying LLM-based analysis and retrieval techniques to produce optimized task-specific prompts (right). This process removes irrelevant instructions while preserving essential context and rules for the specific task scenario.}
    \label{fig:pte}
\end{figure}

\subsection{Test-Time Computation, Reflection \& Validation}
\label{sec:test_time_computation_reflection_and_validation}
Deep Agent implements test-time computation and reflection mechanisms, following recent researches in LLM performance optimization \cite{snell2024scaling}. We ask LLM backend to concurrently generate multiple responses. For simplicity, we apply simpler techniques such as token-level beam search with increased temperature, increase top\_k values and reduced top\_p to promote diversity, or parallel inference runs with different hyperparameter configurations after prefill. Each response undergoes a reflection phase, followed by an integration step that combines the improved responses. Both reflection and integration utilize LLMs with predefined hyperparameters. Through our experiments, we found the test-time computation helped us improve the user experience of the most intricate scenarios, such as prompting C\&D to the user at the right time and only with minimal necessary questions, and such as customizing a large number of options on the UI to exactly the right user specification.

\begin{figure*}[h!] 
    \centering
    \includegraphics[width=1.0\textwidth]{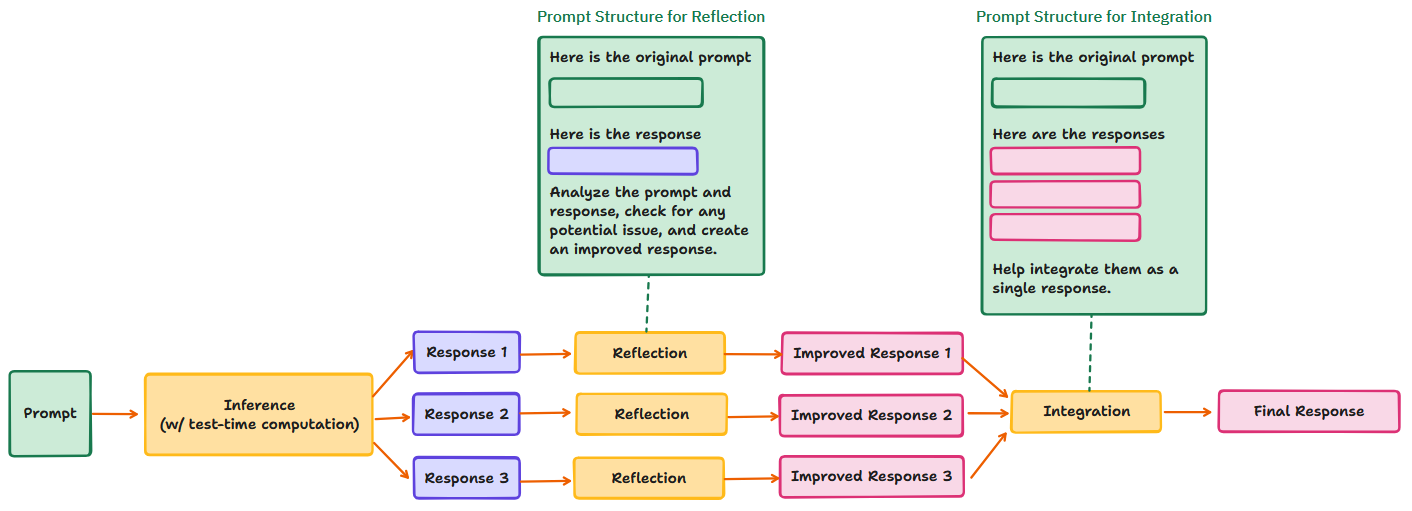} 
    \caption{Starting from a prompt, the system generates multiple responses test-time computation. Each response undergoes a reflection phase for improvement, followed by an integration step that combines the improved responses into a final output. The diagram also roughly illustrates the prompt structures used in reflection and integration.}
    \label{fig:ttc}
\end{figure*}

Deep Agent also incorporates an optional Validator component within the executor to enhance agent system robustness and reliability. The Validator captures snapshots of outcomes after each atomic action, API call, or tool invocation (e.g., screenshots of the UI after actions). These outcome snapshots are analyzed together with the input by more powerful models, which, despite introducing some latency, provide stronger reasoning and in-depth verification. When errors are detected, the Validator halts pending nodes in the task DAG, flagging them for re-planning, and signals the planner to initiate re-planning as needed to consider the impact of the detected error. This particular re-planning with detected error, might temporarily switch to the more powerful model to handle added complexity.

While these mechanisms introduce additional computational overhead, their impact is significantly mitigated by AATC described in Section \ref{sec:aatc}. AATC itself is leveraging Deep Agent to pre-build APIs and tools. These mechanisms substantially improve AATC outcome quality and coverage, converting many runtime inferences into reliable pre-built APIs and tools, and in-turn AATC reduces runtime latency while improving robustness and reliability, enhancing Deep Agent's overall performance.

\begin{figure}[h!]
    \centering
    \includegraphics[width=\columnwidth]{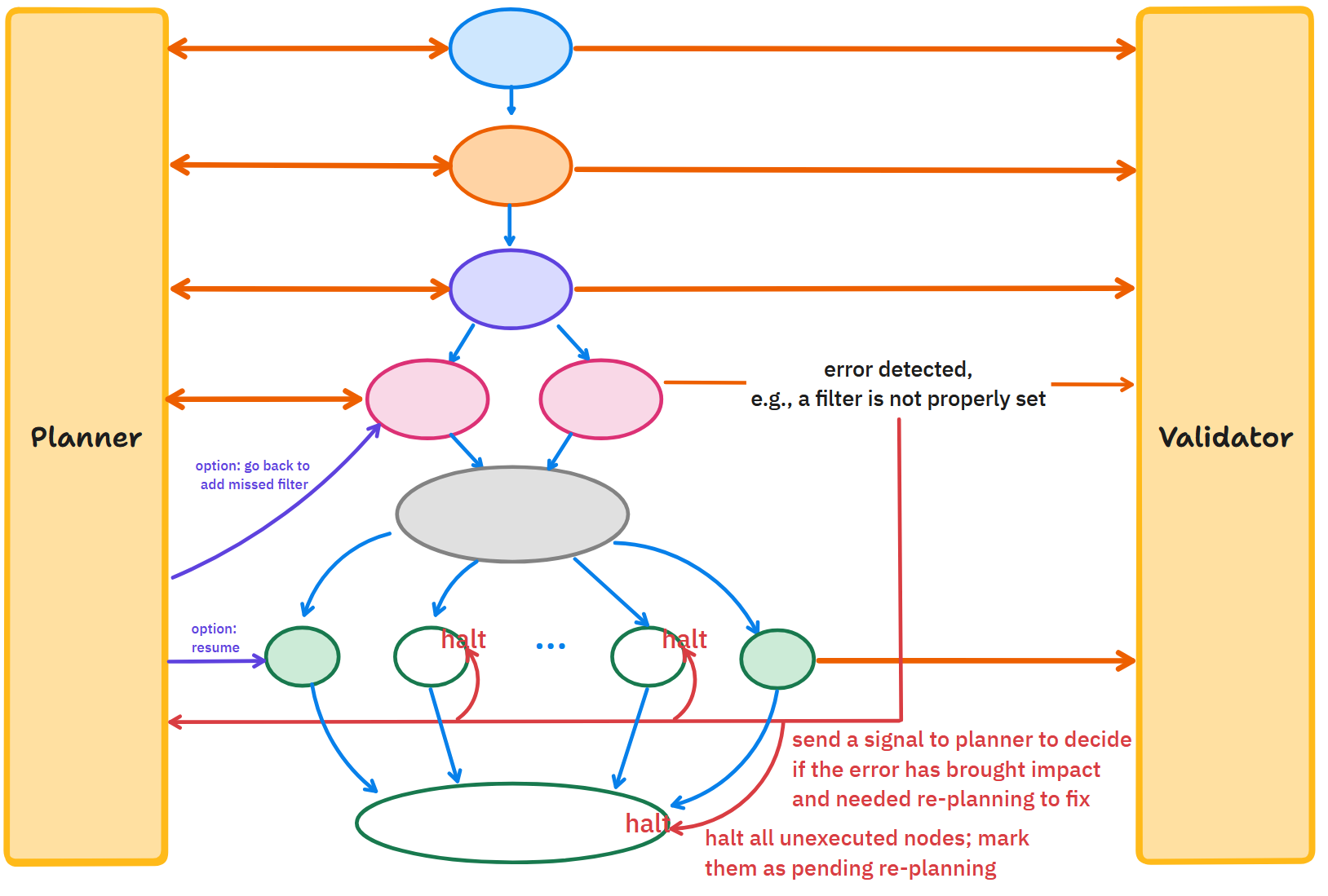}
    \caption{\small Validation architecture showing the interaction between the planner, task DAG and Validator component. We reuse the example from Figure \ref{fig:HTDAG_dynamics} (b-1) here. The Validator monitors execution outcomes at each step as a separate parallel flow. Upon error detection, the Validator halts unexecuted nodes in the task DAG and signals the planner for re-planning. The planner then assesses the error impact: it may proceed if the error is benign (e.g., missed setting "sales" filter but all user-chosen items at the confirmation step were sales items), or it may rescind currently unexecuted nodes and correct the issue (e.g., add missing filter, re-confirm with the user, and restart the item investigation.}
    \label{fig:validator}
\end{figure}

\subsection{Auto Prompt Feedback Learning}
\label{sec:auto_feedback_learning}

Autonomous prompt feedback learning has emerged as a powerful technique to enhance LLM performance without requiring model re-training or fine-tuning \cite{yang2024zhongjing, yilar2024recursive}. Deep Agent implements this approach through a self-instructed feedback workflow that continuously refines prompts based on the agent system's real-world performance and detected errors.

The feedback learning system collects error cases from multiple sources to ensure comprehensive coverage: 1) direct user corrections, 2) reflection disagreements identified during test-time computation, 3) Validator-detected errors during task execution, and 4) additional error cases detected during the autonomous prompt feedback update process (the process requires constant evaluation of prompts, which might reveal new errors). 

\begin{figure*}[h!] 
    \centering
    \includegraphics[width=1.0\textwidth]{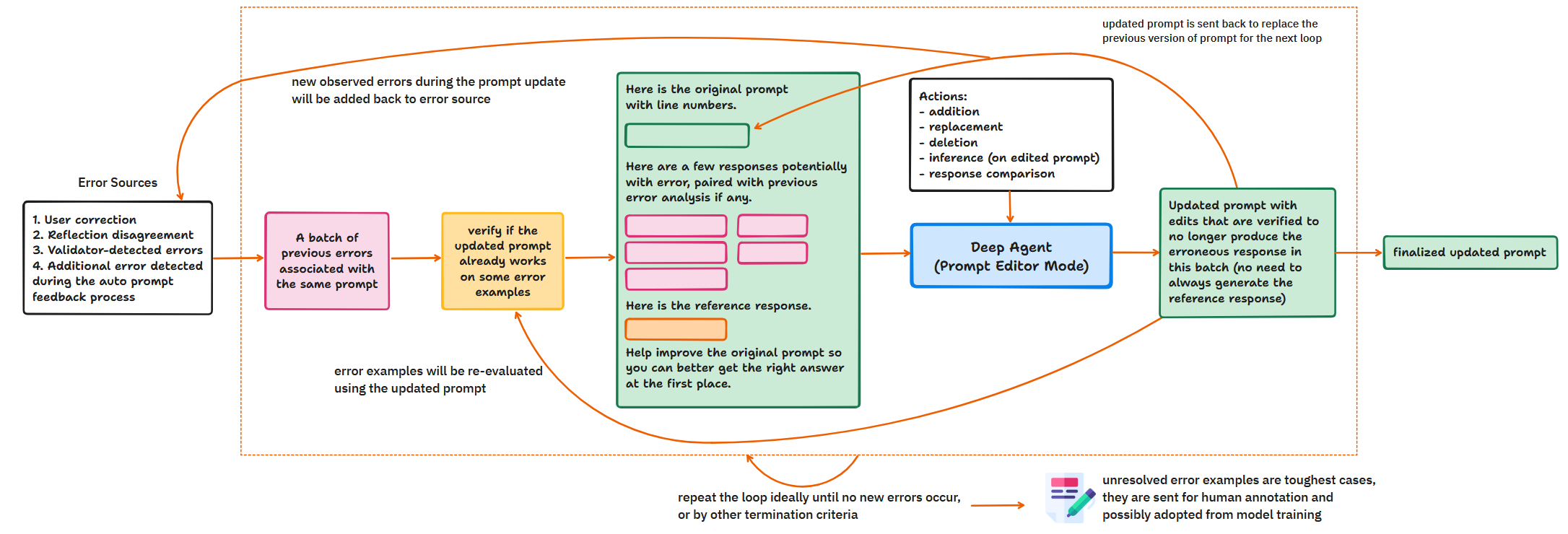} 
    \caption{Autonomous prompt feedback learning pipeline illustrating the closed-loop process for continuous prompt improvement. The system collects error cases from multiple sources (user corrections, reflection disagreements, Validator-detected errors, and automated feedback) and iteratively refines prompts through a systematic process. Each iteration involves evaluating previous errors, verifying prompt updates against error examples, and using Deep Agent in Prompt Editor Mode to make targeted improvements. The process continues until no new errors are detected or other termination criteria are met. Unresolved error cases are flagged for human annotation and potential incorporation into model training. }
    \label{fig:auto_prompt_update}
\end{figure*}

The collected error cases undergo a systematic refinement process, as illustrated in Figure \ref{fig:auto_prompt_update}. Our approach innovates by improving prompts through iterative processing of error batches. The system leverages Deep Agent in ``Prompt Editor Mode'' to analyze each error batch and propose improvements. The following actions or tools are provided to Deep Agent for this process: prompt editing (addition/replacement/deletion), prompt inference, and comparison between inference response and the reference response. Proposed change undergo verifications to ensure we only incorporate prompt changes that address the targeted error cases. It is possible not all error cases in the batch are resolved after testing the proposed prompt changes. We also note the following design details.
\begin{itemize}
    \item \textbf{Precise prompt editing}. Deep Agent's prompts are already structured into distinct sections, and section-wise line numbers are explicitly added to the beginning of each line to further facilitate precise prompt editing.
    \item \textbf{Iterative error reduction}. After each iteration, unresolved error cases are sent back to the error collection. The updated prompt replaces the original version and undergoes testing against all previously successful examples. Any regressions identified during this process are fed back into the error collection for subsequent iterations. Meanwhile, the process also maintains awareness of errors that emerge during the prompt update process itself, feeding these back into the error collection for subsequent iterations. 
    \item \textbf{Termination}. The iterative refinement continues until meeting predefined termination criteria, such as addressing 80\% of error cases or detecting no new errors. Cases that remain unresolved after multiple iterations are flagged for human expert review. These challenging scenarios undergo detailed analysis and may feedback to future model training.
\end{itemize}

We observed that the autonomous feedback process can lead to an accumulation of specific rules in prompts, increasing their size and complexity. This observation  motivated the development of our Prompt Tweaking Engine (PTE) discussed in Section \ref{sec:prompt_tweaking_engine}, which helps manage prompt complexity in runtime while maintaining effectiveness.

Through this comprehensive feedback learning pipeline, Deep Agent achieves continuous refinement of its operational capabilities while maintaining robust performance across diverse scenarios. The system's ability to autonomously identify, address, and validate prompt improvements represents an essential feature for a self-improving AI system.

\bibliographystyle{ACM-Reference-Format}
\bibliography{custom}


\end{document}